# An LP-based hyperparameter optimization model for language modeling

*Amir Hossein Akhavan Rahnama*
*Department of Mathematical Information Technology, University of Jyväskylä,*
*E-mail: amir.h.akhavanrahnama@student.jyu.fi*

*Mehdi Toloo[1]*
*Department of Systems Engineering, Technical University of Ostrava, Ostrava, Czech Republic*
*E-mail: mehdi.toloo@vsb.cz*
*URL: http://homel.vsb.cz/~tol0013/*

*Nezer Jacob Zaidenberg*
*Department of Mathematical Information Technology, University of Jyväskylä,*
*E-mail: nezer.j.zaidenberg@jyu.fi*

**Abstract**

In order to find hyperparameters for a machine learning model, algorithms such as grid search or random search are used over the space of possible values of the models' hyperparameters. These search algorithms opt the solution that minimizes a specific cost function. In language models, perplexity is one of the most popular cost functions. In this study, we propose a fractional nonlinear programming model that finds the optimal perplexity value. The special structure of the model allows us to approximate it by a linear programming model that can be solved using the well-known simplex algorithm. To the best of our knowledge, this is the first attempt to use optimization techniques to find perplexity values in the language modeling literature. We apply our model to find hyperparameters of a language model and compare it to the grid search algorithm. Furthermore, we illustrating that it results in lower perplexity values. We perform this experiment on a real-world dataset from SwiftKey to validate our proposed approach.

---

[1] *Corresponding Author. Postal Address: Sokolska třida 33, 702 00, Ostrava, Czech Republic.  Tel.: +420 597 322 009*

## 1 Introduction

Modern computers can help perform processing, structuring, and managing various kinds of information faster and more efficiently. Nowadays, with the modern popularity of the World Wide Web, larger volumes of data are stored and then processed. This data consists of users activities, transactions etc. Therefore, datasets are growing larger and more complex which led to the concept of Big Data. The Big Data concept is used when data sets have large volume, high velocity and contain immense variety. Before the age of Big Data, explicit programming was the solution to most computing tasks. In the Big Data era, explicit programming is no longer the only solution, especially when complex datasets with high dimension or sparsity are involved.

Samuel (1959) defined Machine Learning as the field of computer science that deals with the computers ability to learn without being explicitly programmed. Statistical Learning Theory is a framework of Machine Learning with the aim of finding a predictive function for a specific dataset (for a more thorough discussion, we refer the reader to Bousquet, Boucheron, & Lugosi , 2004). Halevy, Norvig, & Pereira (2009) showed in a thorough review that the trend in using statistical learning models for language modeling has shown significant improvements over classical linguistic models. Language model is a probability distribution over a sequence of words. According to Goodman (2001), the challenge in language models is the sparsity of data, since many word sequences do not appear in the dataset during the training phase of the model. In a machine learning scenario, like in statistical language models, many hyper-parameters need to be set. It is almost impossible for any machine learning model to be hyper-parameter-free. Bergstra, Bardenet, Bengio, & Kégl (2011) mentioned that in general, the tuning of such hyper-parameters follows a "black-box" approach such as brute force, random or grid search or it relies on expert knowledge. We cannot follow well-defined scientific methods to infer these hyperparameters since these hyper-parameters often cannot be inferred from the data itself. It is apparent that nevertheless, the performance and evaluation of machine learning models depends critically on identifying a good set of hyper-parameters. Furthermore, an automated solution for selecting and optimizing hyper-parameters would drastically reduce the training time of models as well as the exhaustive feature engineering efforts. More recently, the tuning of hyper-parameters is solved as a non-convex optimization problem and is carried out in a systematic manner. We prefer systematic approaches since the former approaches have not scaled well. For example, Bergstra & Bengio (2012) highlighted that random search is a common method when using Artificial Neural Networks. On the other hand, Bergstra et al. (2011) pointed out that models such as Support Vector Machine (SVM) which are commonly used for convex optimization excel with grid search.

*1.1    Language model*

Statistical Language models are probability distributions of words sequence. Any sequence of $n$ items ($w_1 \ldots w_{n-1} w_n$) is called an *n*-gram:

$$P(w_1 \ldots w_i) = P(w_1) \times P(w_2|w_1) \times \ldots \times P(w_i|w_1 \ldots w_{i-1}) \qquad (1)$$

In calculating the probability of a given sequence, Markov Chain rule is used to simplify computations for higher values of $i$ in formula (1). For example, depending on the context of the problem we are solving, we can assume that the probability of any sequence depends only on the last two words preceding it (*trigram* condition):

$$P(w_i|w_1 \ldots w_{i-1}) \approx P(w_i|w_{i-2} w_{i-1}) \approx \frac{C(w_{i-2} w_{i-1} w_i)}{C(w_{i-1} w_i)} \qquad (2)$$

Although it seems that formula (2) is straightforward, in order to be able to predict sequences that have occurred in the test data but not in the training set, we use smoothing algorithms. In such a case, Chen & Rosenfeld (1999) showed that different smoothing algorithms can be used, however in our study, we opt the interpolation smoothing algorithms among the rest due to their flexibility and computational efficiency. Brants et al. (2007) suggested using "Stupid Backoff" method that is non-stochastic in contrast to other interpolation smoothing algorithms. Goodman (2001) illustrated that stochastic smoothing algorithm such as the interpolation algorithm generates probabilistic scores that are both easily interpretable and very efficient. We formulate the interpolation algorithm for general case of *n*-gram which is based on the representation from Goodman (2001):

$$\begin{aligned} P(w_i|w_1 \ldots w_{i-1}) &= \lambda_n P(w_i|w_{i-n+1} \ldots w_{i-1}) + \cdots + \lambda_2 P(w_i|w_{i-1}) + \lambda_1 P(w_i) \\ &= \sum_{j=2}^{n} \lambda_j P(w_i|w_{i-j+1} \ldots w_{i-1}) + \lambda_1 P(w_i) \end{aligned} \qquad (3)$$

where $\sum_{j=1}^{i-n+1} \lambda_j = 1$. In other words, it is assumed that $P(w_i|w_1 \ldots w_{i-1})$ is a convex combination of $P(w_i|w_{i-j+1} \ldots w_{i-1})$ for $j = 1, \ldots, (i-n+1)$. It should be noted that because $P(w_i|w_n \ldots w_{i-1}) \in [0,1]$ then $P(w_i|w_1 \ldots w_{i-1}) \in [0,1]$. Since calculating $P(w_i|w_1 \ldots w_{i-1})$ for large $i$ can become comptutianally challenging in large-scale data, the value of $j$ can be fixed.

Brown, Desouza, Mercer, Pietra, & Lai (1984) pointed out that one of the common measures to find all hyperparameters $\lambda_i \geq 0 \; \forall i$. The following is a representation of the measure called *perplexity*:

$$\left(\frac{1}{\prod_{i=1}^{N} P(w_i | w_1 \ldots w_{i-1})}\right)^{1/N} \tag{4}$$

*1.2  Optimization*

Operations research is concerned with important concepts such as efficiency and scarcity, i.e. the optimal allocation of resources and shortage of resources. Optimization problems deals with maximizing or minimizing a specific quantity called the *objective function* that depends on a set of input variables. Input variables can be both independent or dependent on one another via constraints. Optimization problems have a number of *constraints* that they have to satisfy. If possible solutions satisfy all constraint in an optimization problem, they become *feasible solutions*. An *optimal* solution is a feasible solution that maximizes or minimizes the objective function. As a matter of fact, solving an optimization problem is equivalent to using methods to find the optimal solutions.

Linear programming is a special case of an optimization problem of a linear objective function, whether maximization or minimization, subject to linear equality and inequality constraints. To formulate an optimization problem as a linear programming problem, four main assumptions needs to be considered: proportionality, additivity, divisibility and deterministic. Since 1947 when George Dantzig developed the simplex method, linear programming has been utilizing extensively in economics, management, transportation, energy, telecommunications, and manufacturing and computer science. It is of utmost importance to mention that solving a linear programming problem is easier than other optimization problem types such as nonlinear or integer programming problems. A nonlinear programming problem is when some of the constraints or the objective function are nonlinear while an integer programming problem is a case where variables need to be integers. Both of these types of problems are NP-hard in many cases, whereas linear programming problems are solved efficiently even in worst cases. For a deeper discussion about linear and non-linear programming problems, we refer the reader to Bazaraa, Jarvis, & Sherali (2010) and Bazaraa, Sherali, & Shetty (2013).

*1.3 Monte Carlo Simulation*

Computer simulations are set of programs that run during a time interval to capture the behavior of the model. Simulation is not an optimization technique; however, it ranks high among the most used techniques in statistics and operations research. Simulation is a technique that determines measure of performance of stochastic systems by generating and recording occurrences of different events as if they were actually happening and measuring model's output. Simulation is a key tool in analyzing complex models that are not easily solvable with mathematical programming. Hillier & Lieberman (2001) emphasized that mathematical models give a far better insight into cause and effects of the model and they are superior to measures of a simulation process, but many real-world problems are too complex to be mathematically modeled. Monte Carlo experiments are type of computer simulations that are based on random samplings of each variable with many possible outcomes. Vose (2008) indicated that under the assumption of the law of large numbers and using Markov Chains and Monte Carlo sampler, these experiments can solve problems with probabilistic interpretations under Markov Chain assumptions as well.

## 2   Proposed Approach

As mentioned before, an interpolated smoothing algorithm for n-gram models is as follows:

$$P(w_i|w_1 \ldots w_{i-1}) = \sum_{j=1}^{n} \lambda_j P_{j-gram}(w_i|w_1 \ldots w_{i-1}) \qquad (5)$$

where $\sum_{i=1}^{n} \lambda_i = 1$ and $\lambda_i \geq 0 \; \forall i$. As a matter of fact, $P(w_i|w_1 \ldots w_{i-1})$ is a convex combination of $P_{1-gram}(w_1), P_{j-gram}(w_i|w_1 \ldots w_{i-1})$

```
j = 0;
for ( λ₁ = 0; to λ₁ <= 1; step_{λ₁}++)
    for ( λ₂ = 0; to λ₂ <= 1 − λ₁ ; step_{λ₂}++)
        ⋮
        for ( λ_{n-1} = 0; to λ_{n-1} <= 1 − ∑_{j=1}^{n-2} λ_j ; step_{λ_{n-1}}++)
            {P(j) = perplexity(λ₁, …, λ_{n-1}, 1 − ∑_{j=1}^{n-1} λ_j );
             j + +;
            }
return min(P(j))
```

It is clear on inspection that the above algorithm terminates after checking all feasible combinations of the hyperparamter space within a defined step size and returns the minimum perplexity value. On one hand, selection of smaller step sizes increases the accuracy of the minimum value returned by the aforementioned algorithm which is rather important. On the other hand, it significantly increases the required completion time in an exponential form. The problem shall become even more complex if one selects a higher order n-gram.

We formulate the following fractional nonlinear programming with the aim of finding the optimal perplexity value from all probability of terms in the training set:

$$\min \left(\frac{1}{\prod_{i=1}^{N} P(w_i|w_1 \ldots w_{i-1})}\right)^{1/N}$$
$$\text{s.t.}$$
$$\sum_{j=1}^{n} \lambda_{n-j+1} P_{(n-j+1)gram}(w_i|w_1 \ldots w_{n-j}) - P(w_i|w_1 \ldots w_{i-1}) = 0 \quad i = 1, \ldots, N \quad (6)$$
$$\sum_{j=1}^{n} \lambda_j = 1$$
$$\lambda_j \geq 0 \quad\quad\quad\quad\quad\quad\quad\quad\quad\quad\quad\quad\quad\quad\quad\quad\quad\quad j = 1, \ldots, n$$

where $\lambda_j$ for $j = 1, \ldots, n$ and $P(w_i|w_1 \ldots w_{i-1})$ for $i = 1, \ldots, N$ are decision variables and $P_{(n-j+1)gram}(j = 1, \ldots, n)$ and $(1,1,\ldots,1)$ are technological coefficients which form the following constraint matrix in $\mathbb{R}^{(N+1) \times (N+n)}$ space:

$$\begin{bmatrix} P_{(n)gram}(w_1) & P_{(n-1)gram}(w_1) & \cdots & P_{(1)gram}(w_1) & -1 & 0 & \cdots & 0 \\ P_{(n)gram}(w_2|w_1) & P_{(n-1)gram}(w_2|w_1) & \cdots & P_{(1)gram}(w_2) & 0 & -1 & \cdots & 0 \\ \vdots & \vdots & \ddots & \vdots & \vdots & \vdots & \ddots & \vdots \\ P_{(n)gram}(w_N|w_1 \ldots w_{n-1}) & P_{(n-1)gram}(w_N|w_1 \ldots w_{n-2}) & \cdots & P_{(1)gram}(w_N) & 0 & 0 & \cdots & -1 \\ 1 & 1 & \cdots & 1 & 0 & 0 & \cdots & 0 \end{bmatrix}$$

Model (6) is a fractional non-linear programming due to perplexity being a nonlinear and a fractional measure. The model takes all convex combinations of probabilities of all *n*-grams into consideration in order to choose the minimum perplexity, however employing Charnes & Cooper's (1962) transformation formula, Model (6) is equivalent to the following non-linear problem:

$$\max \prod_{i=1}^{N} P(w_i|w_1 \ldots w_{i-1})$$
$$\text{s.t.}$$
$$P(w_i|w_1 \ldots w_{i-1}) = \sum_{j=1}^{n} \lambda_{n-j+1} P_{(n-j+1)gram}(w_i|w_1 \ldots w_{n-j}) \quad i = 1, \ldots, N \quad (7)$$
$$\sum_{j=1}^{n} \lambda_j = 1$$
$$\lambda_j \geq 0 \quad\quad\quad\quad\quad\quad\quad\quad\quad\quad\quad\quad\quad\quad\quad\quad\quad\quad j = 1, \ldots, n$$

On the one hand, Model (7) has tackled the fractional aspects of Model (6); on the other hand, it is a non-linear model. Generally, non-linear functions have unstable behaviors and can have multiple local optimal solutions. There are no general algorithms to solve non-linear problems i.e. to reach the global optimal solution. Therefore linearization of non-linear functions has gotten the wide attention of the optimization research community. This transformation will ensure that the global optimal solution will be reached at optimality. An alternative approach is to use a suitable linear approximation technique that guarantees an acceptable estimated optimal solution. Model (7) has a special structure that helps us to find a linear approximation model: (i) its feasible region is a polyhedron set, (ii) its non-linearity is just because of its objective function, (iii) objective function is a multiplication of a finite number of probabilities which are between 0 and 1. Let $f(x_1, ..., x_n) = \prod_{i=1}^{n} x_i$, $g(x_1, ..., x_n) = \sum_{i=1}^{n} x_i$, and $\forall i, \ 0 \leq x_i \leq 1$. It is easy to verify that the linear function $g(\boldsymbol{x})$ is a suitable apporoximation of the nonliner function $f(\boldsymbol{x})$ where $\boldsymbol{x} = (x_1, ..., x_n)$. For a deeper disucssion on approximation of non-linear convex functions, see Boyd & Vandenberghe (2004). Figure 1 illustrtaes the fact that difference values between geometric function $f(x_1, x_2)$ and linear function $g(x_1, x_2)$ is nelgibible.

As a result, since the objective function of Model (7) is a product of $N$ probability sequences which are within [0,1] we suggest the following approximation LP Model (8) that maximizes the sum of $P(w_i | w_1 ... w_{i-1})$ over the same feasible region of Model (7).

$$\max \sum_{i=1}^{N} P(w_i | w_1 ... w_{i-1})$$
$$\text{s.t.}$$
$$P(w_i | w_1 ... w_{i-1}) = \sum_{j=2}^{n} \lambda_j P(w_i | w_{i-j+1} ... w_{i-1}) + \lambda_1 P(w_i) \quad i = 1, ..., N \quad (8)$$
$$\sum_{j=1}^{n} \lambda_j = 1$$
$$\lambda_j \geq 0 \quad\quad\quad\quad\quad\quad\quad\quad\quad\quad\quad\quad\quad\quad j = 1, ..., n$$

The above model has the computational advantage of being a linear programming model i.e. both the objective function and the constraints are linear. Referencing to Bazaraa et al. (2013), using the simplex method to solve Model (8) needs on the order of the $N$ to $3N$ iterations. Each iteration involves $n(N + 2)$ multiplications and $(N + 1)(n - 3)$ additions. Because the number of operation for each iteration is of order $O(N^2)$ and therefore the average empirical complexity of the simplex method is $O(N^3)$. As a result, using proposed LP Model (8) significantly reduced both the complexity and computational burden of the NLP Model (6).

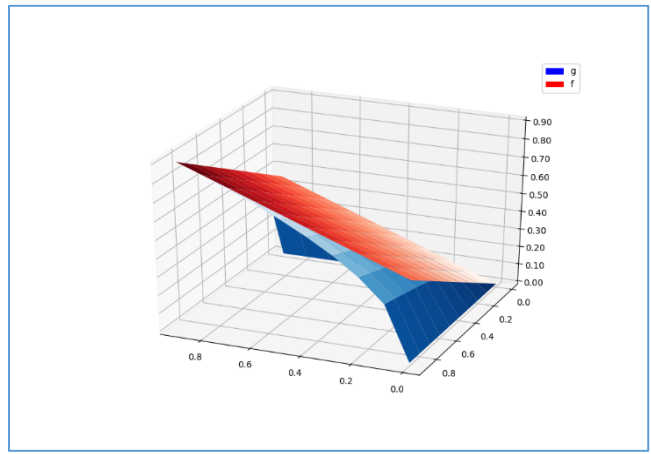

*Figure 1 - Surface Plot of Perplexity and Arithmetic Mean*

## 2.1 Experiments: 3-gram models with SwiftKey

Goodman (2001) and Halevy et al. (2009) have shown that trigram models have proven to be sufficient and accurate for multiple cases. Therefore, we restrict formula (3) for trigram condition throughout this study:

$$P(w_i|w_{i-2}w_{i-1}) = \lambda P_{3-gram}(w_i|w_{i-2}w_{i-1}) + (1-\lambda)[\mu P_{2-gram}(w_i|w_{i-1}) + (1-\mu)P_{1gram}(w_i)]$$

where $\lambda$ and $\mu$ are constants such that $\lambda, \mu \in [0,1]$. Moreover, $P_{3-gram}(w_i|w_{i-2}w_{i-1})$, $P_{2-gram}(w_i|w_{i-1})$, and $P_{1-gram}(w_i)$ are between 0 and 1, which follows $P(w_i|w_{i-2}w_{i-1}) \in [0,1]$. Different values of $\lambda$ and $\mu$ will result in different values for $P(w_i|w_1 \ldots w_{i-1})$. In the above formula if $\lambda = 0$ and $\mu = 1$, then $P(w_i|w_1 \ldots w_{i+1}) = 0$ which leads to an undefined perpexity (weak model condition). It should be remarked that the formulations of 3-grams might differ in different literature. In formula (3), we follow the notation in Halevy et al. (2009). As inspection makes clear, it can be rewritten as bellow:

$$P(w_i|w_{i-2}w_{i-1}) = \lambda_3 P_{3-gram}(w_i|w_{i-2}w_{i-1}) + \lambda_2 P_{2-gram}(w_i|w_{i-1}) + \lambda_1 P_{1-gram}(w_i)$$

where $\lambda_1 + \lambda_2 + \lambda_3 = 1$ and $\lambda_1, \lambda_2, \lambda_3 \geq 0$.

As result, Model (8) will be equivalent to the following form, when written for the special case of 3-grams:

$$\max \sum_{i=1}^{N} P(w_i|w_1 \ldots w_{i-1})$$
$$\text{s.t.}$$
$$P(w_i|w_1 \ldots w_{i-1}) = \lambda_3 P_{3-gram}(w_i|w_{i-2}w_{i-1}) + \lambda_2 P_{2-gram}(w_i|w_{i-1}) + \lambda_1 P_{1-gram}(w_i) \quad i = 1, \ldots, N \quad (9)$$
$$\lambda_1 + \lambda_2 + \lambda_3 = 1$$
$$\lambda_1, \lambda_2, \lambda_3 \geq 0$$

## 2.2 Data and Result

In this case study, we apply our model on a dataset published by SwiftKey[2]. The dataset includes 2360148 English sentences in total. The interesting part of SwiftKey dataset is that sentences are given from three main categories: blogs posts, news websites, and Twitter. Since the human language can vary in both vocabulary and structure in different context, SwiftKey dataset acts as a good diverse summary of those different contexts which has a built-in complexity that reflects a real-world scenario in which a language model needs to predict. In order to test our model, we follow the cross-validation strategy which is the de-facto standard of training machine learning models. In this study, we formed our cross-validation set with 60% of training and the rest equally divided between validation and test sets.

---

[2] SwiftKey is an input method that employs various artificial intelligence technologies to predict the next word the user intends to type. For an available free download, visit: https://swiftkey.com/en.

| Data | Number of sentences |
|---|---|
| Each fold | 73754 |
| Total | 2360148 |

Table 1 - Cross Validation Size

In order to avoid sparsity, we set the minimum thresholds for our n-grams ($n = 1, 2, 3$). We only include 1-grams with threshold of 19 occurences, 2-grams with 29 occurences and 3-grams with 39. After this, the size of each n-grams is as follows:

| n-gram | Size |
|---|---|
| 1-gram | 84249 |
| 2-gram | 374498 |
| 3-gram | 216029 |

Table 2 - Size of N-grams

We then performed the k-fold cross-validation with $k = 32$. In each interation, we excluded one set as testset and then we ran the experiement 32 times. Each time sampling a new training, validation and test set. In each experiment, we have recorded the perplexity that our model achieved as **Perplexity**$^{LP}$ and we then compared the result to the grid-search which was denoted by **Perplexity**$^{GS}$. In our experiment, we obtained $\lambda_3^{LP} = 0.1, \lambda_2^{LP} = 0,8001, \lambda_1^{LP} = 0.0999$, that resulted in **Perplexity**$^{LP} = 69.2605$ while $\lambda_3^{GS} = 0.1, \lambda_2^{GS} = 0.09, \lambda_1^{GS} = 0.81$ which resulted in **Perplexity**$^{GS} = 93.5025$ on average throughtout our 32-fold validation set. The result of **Perplexity**$^{GS}$ was the best between choosing 0.1 and 0.01 as its step-size (see Table 3).

|  | Perplexity ||
|---|---|---|
|  | Expected Value (EV) | Standard Deviation (SD) |
| **LP-based Model** | 69.2605 | 179.196 |
| **Grid Search** | 93.5025 | 229.231 |

Table 3 - Comparing EV and SD

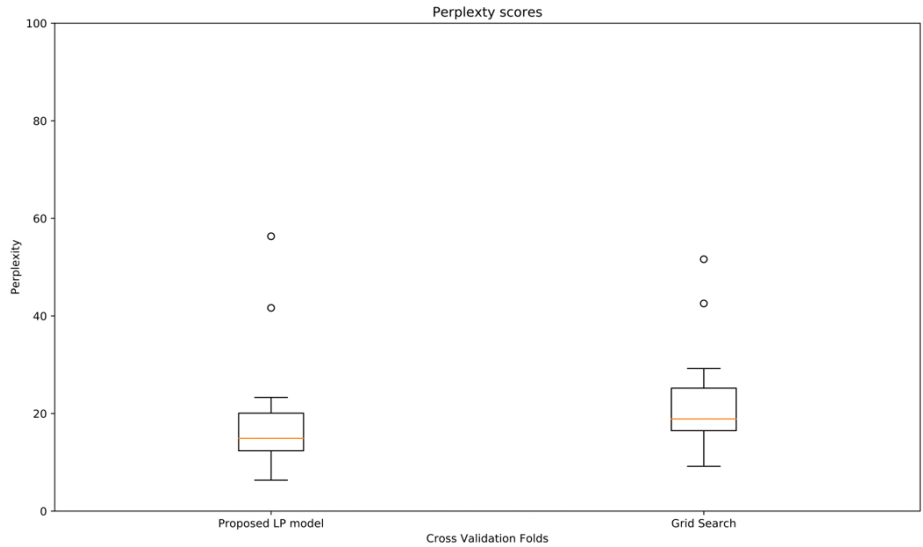

*Figure 2 - Boxplot of k-fold cross validation result*

As we can see in Figure 2, our proposed method clearly out-performs the grid search in both average perplexity values across folds. Furthermore, our method is much more resistant across different validation folds. Note that in order to have a clear visualization, we have omitted the outliers, however, they were included in the calculation of expected values.

## 3   Conclusion and Remarks

In this study, we have considered the problem of finding hyperparameters for a linear smoothing algorithm as a constrained optimization problem. Due to the complexity of using perplexity as an objective function in our optimization model, we have proposed a linear approximation to perplexity. With the new proposed measure, we next suggested an LP model that can find the most optimal solution to hyperparamters of a 3-gram smoothed language model. After that, we showed that our model can outperform grid search algorithm using the SwiftKey dataset.

We believe that our model can be applied to the state-of-the-art language model, namely Neural Probabilist Language Model of Bengio, Ducharme, Vincent, & Janvin (2003). In addition to that, we propose that our model can be used with Mikolov, Karafiát, Burget, & Khudanpur's (2010) Recurrent Neural Network based language model. While our approach mainly aims at tackling the problem of language modeling, we believe that by our proposed approach is useful in general machine learning settings as well especially in the case where large-scale datasets are used.

**Acknowledgements**

Mehdi Toloo would like to thank the European Social Fund (CZ.1.07/2.3.00/20.0296) and the Czech Science Foundation (GAČR 17-23495S).

**References**

Bazaraa, M. S., Jarvis, J. J., & Sherali, H. D. (2010). *Linear programming and network flows* (4th ed.). John Wiley & Sons.

Bazaraa, M. S., Sherali, H. D., & Shetty, C. M. (2013). *Nonlinear Programming: Theory and Algorithms*. John Wiley & Sons.

Bengio, Y., Ducharme, R., Vincent, P., & Janvin, C. (2003). A Neural Probabilistic Language Model. *The Journal of Machine Learning Research*, *3*.

Bergstra, J., Bardenet, R., Bengio, Y., & Kégl, B. (2011). Algorithms for Hyper-Parameter Optimization. *Advances in Neural Information Processing Systems*.

Bergstra, J., & Bengio, Y. (2012). Random Search for Hyper-Parameter Optimization. *Journal of Machine Learning Research*, *13*(1).

Bousquet, O., Boucheron, S., & Lugosi, G. (2004). Introduction to Statistical Learning Theory. In *Advanced Lectures on Machine Learning* (pp. 169–207). Springer Berlin Heidelberg.

Boyd, S. P., & Vandenberghe, L. (2004). *Convex optimization*. Cambridge University Press.

Brants, T., Brants, T., Popat, A. C., Xu, P., Och, F. J., & Dean, J. (2007). Large language models in machine translation. In *The joint conference on empirical methods in natural language processing and computational natural language learning* (pp. 858--867).

Brown, P. F., Desouza, P. V., Mercer, R. L., Pietra, V. J. D., & Lai, J. C. (1984). *Class Base n-gram models of natural language*. *Computational Linguistics* (Vol. 18). Association for Computational Linguistics.

Charnes, A., & Cooper, W. W. (1962). Programming with linear fractional functionals. *Naval Research Logistics Quarterly*, *9*.

Chen, S. F., & Rosenfeld, R. (1999). A Gaussian Prior for Smoothing Maximum Entropy Models. *School of Computer Science, Carnegie Mellon University*.

Goodman, J. (2001). A Bit of Progress in Language Modeling. *Technical Report*.

Halevy, A., Norvig, P., & Pereira, F. (2009). The Unreasonable Effectiveness of Data. *IEEE Intelligent Systems*, *24*(2).

Hillier, F. S., & Lieberman, G. J. (2001). *Introduction to operations research*. McGraw-Hill.

Mikolov, T., Karafiát, M., Burget, L., & Khudanpur, S. (2010). Recurrent neural network based language model. *Interspeech*, *2*.

Samuel, A. L. (1959). Some Studies in Machine Learning Using the Game of Checkers. *IBM Journal of Research and Development*, *3*(3).

Vose, D. (2008). *Risk analysis : a quantitative guide* (3rd ed.). Wiley.